\ifdefined\XeTeXversion
\else
  \pdfoutput=1
\fi
\PassOptionsToPackage{table}{xcolor}
\documentclass[10pt,logo,onecolumn,copyright]{nv}

\usepackage{graphicx}
\usepackage{xcolor}
\usepackage[utf8]{inputenc}
\usepackage[T1]{fontenc}
\usepackage{times}
\usepackage{amsmath}
\usepackage{amssymb}
\usepackage{mathtools}
\usepackage{booktabs}
\usepackage{multirow}
\usepackage{tabularx}
\usepackage{subcaption}
\usepackage{float}
\usepackage{url}
\usepackage[nameinlink]{cleveref}
\usepackage[square,sort,comma,numbers]{natbib}
\usepackage{xspace}

\definecolor{nvidiagreen}{HTML}{76B900}
\definecolor{nvgreen}{HTML}{D3E9AD}
\definecolor{nvgreenlight}{HTML}{EAF5D8}
\definecolor{nvgray}{HTML}{5B5E5F}
\hypersetup{
  colorlinks=true,
  citecolor=nvidiagreen,
  linkcolor=nvidiagreen,
  urlcolor=nvidiagreen
}
\crefname{section}{Sec.}{Secs.}
\crefname{figure}{Figure}{Figures}
\crefname{table}{Table}{Tables}
\Crefname{figure}{Figure}{Figures}
\Crefname{table}{Table}{Tables}
\renewcommand{\today}{2026-08-17}

\newcommand{\solpi}{SoL-Pi\xspace}
\newcommand{\edgebench}{EdgeBench\xspace}
\newcommand{\actionfusion}{Action Fusion\xspace}
\newcommand{\onlinecompact}{Online Context Compact\xspace}
\newcommand{\observationpack}{ObservationPack\xspace}
\newcommand{\lunaepdr}{Evidence-Preserving Reducer\xspace}

\title{\solpi: Recursively Scaling Auto-Research Loops for Efficient Agent Harness}

\author{
\parbox{\linewidth}{
\centering
\vspace{-5pt}
{\fontsize{9pt}{16pt}\selectfont\textbf{Haozhe Liu\textsuperscript{1\thinspace*}, ~ Tian Ye\textsuperscript{1\thinspace*}, ~ Sensen Gao\textsuperscript{2\thinspace\textdagger}, ~ Qihang Cao\textsuperscript{2\thinspace\textdagger}, ~ Yitong Li\textsuperscript{1\thinspace\textdagger}, ~ Mingchen Zhuge}}
\\
\vspace{0.35em}
{\fontsize{9pt}{16pt}\selectfont\textbf{Duomin Wang\textsuperscript{1}, ~ Ruihua Zhang\textsuperscript{1}, ~ Ping Luo\textsuperscript{1}, ~ Jiawang Bian\textsuperscript{2}, ~ Lei Zhu\textsuperscript{1}, ~ Ligeng Zhu\textsuperscript{1}, ~ Enze Xie\textsuperscript{1}, ~ Song Han\textsuperscript{1\thinspace3}}}
\\
\vspace{2.2mm}
{\fontsize{8.7pt}{11pt}\selectfont \textsuperscript{1}NVIDIA \quad \textsuperscript{2}NTU \quad \textsuperscript{3}MIT}
\\
\vspace{0.2em}
{\fontsize{8.2pt}{10pt}\selectfont \textsuperscript{*} Equal contribution. \quad \textsuperscript{\textdagger} Core contributors.}
\\
\vspace{5pt}
{\fontsize{9pt}{11pt}\selectfont
\href{https://github.com/NVlabs/SoL-Pi}{\raisebox{-0.15em}{\includegraphics[height=1em]{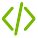}}\hspace{0.35em}\textbf{Code}}
\hspace{1.8em}
\href{https://nvlabs.github.io/SoL-Pi/}{\raisebox{-0.15em}{\includegraphics[height=1em]{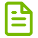}}\hspace{0.35em}\textbf{Blog Post}}}
\\
\vspace{2pt}
}
}

\begin{abstract}
As coding agents move from supervised code completion to unattended, around-the-clock exploration, their work expands from isolated predictions into long trajectories of reasoning, tool use, and feedback. Token efficiency therefore becomes important for scaling recursive self-improvement. We take an RSI-inspired approach at the harness layer, scaling auto-research loops across increasingly numerous and diverse environments for harness rollouts. At this scale, the process yields reusable improvements that transfer beyond their development setting, moving automated harness discovery toward production-level outcomes. Four mechanisms survive selection and form SoL-Pi, spanning action execution, context compaction, observation handling, and delegated reading. On the 51-task EdgeBench evaluation, SoL-Pi achieves performance comparable to Pi across GPT-5.6 Sol and Opus 5 while reducing recorded token traffic by 44.7--49.0\% and API cost by about one third. In other words, estimated hourly savings are \$8.75--\$13.50 relative to native Codex and Claude Code harnesses, and \$4.36--\$5.71 relative to Pi.

\end{abstract}

\begin{document}

\maketitle

\vspace{1pt}

\ifdefined\SoLPiTeaserBeforeAbstract
\else
  \begin{figure}[h]
    \centering
    \includegraphics[width=\linewidth]{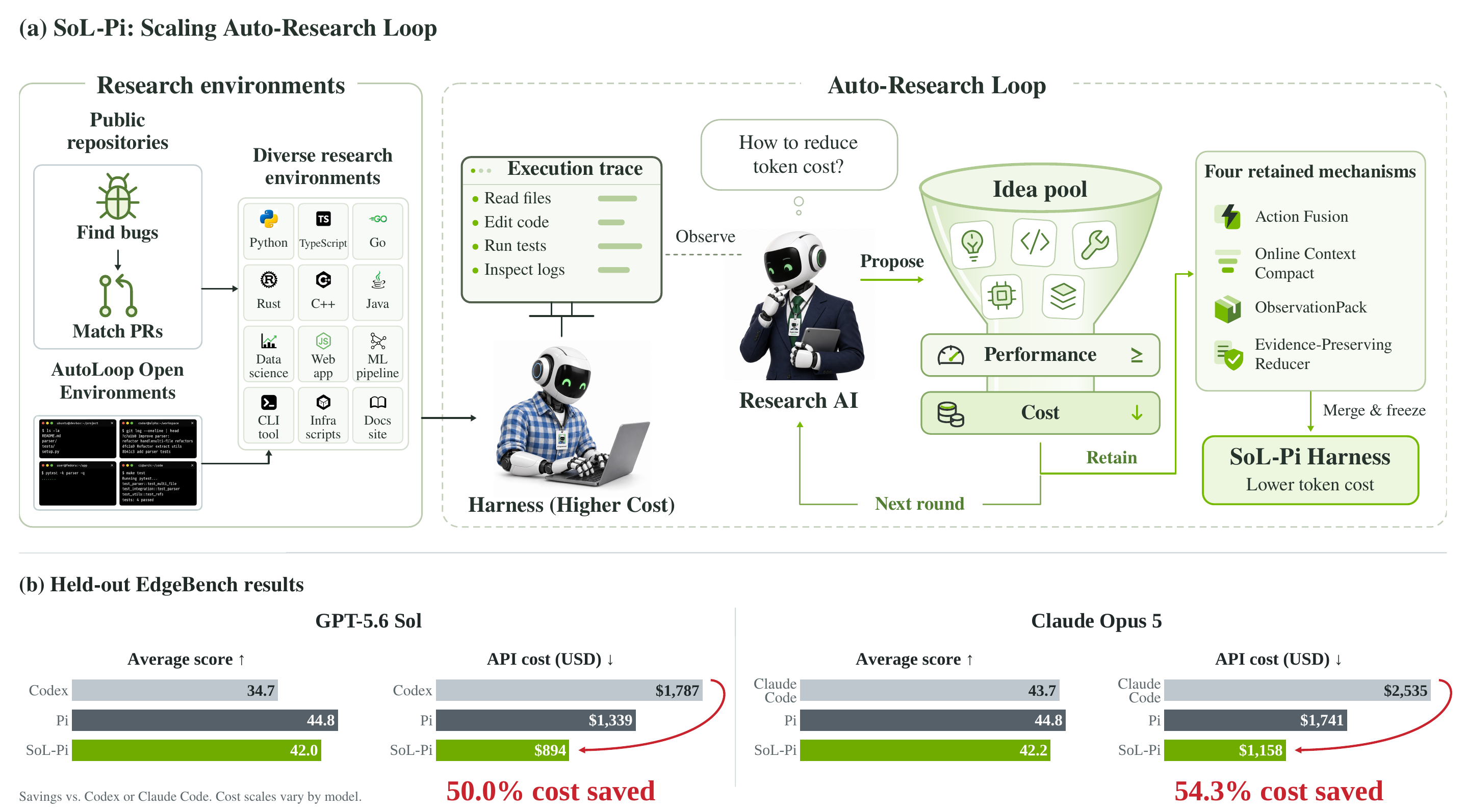}
    \vspace{-5pt}
    \caption{\textbf{SoL-Pi discovers a more token-efficient harness through automated research.}
    (a) \textbf{SoL-Pi: Scaling Auto-Research Loop.} Prepared research environments supply tasks to an AI running the base harness. A research AI inspects its execution traces, proposes candidate changes, and filters the idea pool through capability and efficiency gates. Four retained mechanisms are integrated and refined into SoL-Pi before the harness is frozen for evaluation on unseen benchmarks. The trace, ideas, and gate symbols are schematic: capability is checked within fixed tolerances, and held-out results never feed back into search.
    (b) Example results on EdgeBench: average score and API cost for the native harnesses (Codex with GPT-5.6 Sol and Claude Code with Opus 5), Pi, and the complete four-mechanism SoL-Pi harness. SoL-Pi reduces API cost by \textbf{50.0\%} relative to Codex on GPT-5.6 Sol and by \textbf{54.3\%} relative to Claude Code on Opus 5. }
    \label{fig:teaser}
\end{figure}

\fi

\Needspace{12\baselineskip}
\section{Introduction}
\label{sec:introduction}

Advances in foundation models enable agents to tackle increasingly open-ended tasks over longer horizons with less supervision~\cite{chen2021codex,yang2024sweagent,kwa2025measuring,cursor2026selfdriving}. This shift supports applications such as autonomous research, software engineering agents, self-evolving personal assistants, and early forms of recursive self-improvement (RSI)~\cite{openai2026gpt53codex,aiscientist_v2,zhang2026dgm,zhang2026hyperagents}. As agents operate over longer horizons, task-level token efficiency becomes a first-order systems concern~\cite{anthropic2025longrunningharness,ye2025agentfold}. Existing efficiency work has primarily focused on lowering the cost per token through faster attention kernels and serving infrastructure~\cite{dao2024flashattention,kwon2023vllm}, model compression techniques such as quantization~\cite{xiao2023smoothquant,frantar2022gptq}, or the use of cheaper models~\cite{chen2023frugalgpt,ong2024routellm}. In this paper, we explore an orthogonal direction: improving token use through the agent harness that mediates interactions between the model and its environment.

Harness-level optimization can improve efficiency without additional model training, complementing infrastructure- and model-level approaches~\cite{lee2026metaharness,lin2026ahe}. However, optimizing a harness is difficult in practice. Since tool use, context management, verification, delegation, recovery, and termination are tightly coupled, a change that is locally beneficial may cause downstream failures or shift token costs to later stages of execution. In practice, harness development often requires substantial human effort to inspect long execution traces, identify recurring failure modes, and translate these observations into code changes. This process is costly and difficult to scale across tasks and environments.

To accelerate this process, we adopt an RSI-inspired approach in which an AI optimizer iteratively improves the agent harness for token efficiency. In the \emph{SoL-Pi: Scaling Auto-Research Loop} workflow (\cref{fig:teaser}(a)), the research AI observes execution traces from a separate agent running the base harness, proposes candidate changes, and tests them in prepared research environments. Capability and efficiency checks determine which candidates are retained, while development results guide subsequent iterations.

Recent work has demonstrated the feasibility of automated harness improvement. Meta-Harness searches over executable harness programs and evaluates their transfer to held-out datasets and models~\cite{lee2026metaharness}, while Recursive Harness Self-Improvement (RHI) iteratively refines prompt-level specifications of the agent loop for individual tasks~\cite{lee2026rhi}. However, a recent study using held-out tasks finds that evolved harnesses can overfit the tasks used during search and provide only marginal gains on unseen tasks \cite{wang2026rethinkingharnessevolution}. These findings motivate a clear separation between search feedback and final evaluation \cite{pmlr-v202-gao23h}. To discover transferable efficiency improvements, we introduce SoL-Pi, a system for discovering harness improvements that transfer beyond the tasks used during search. SoL-Pi organizes autonomous research as a broad-to-deep funnel that separates candidate development from held-out validation and scales through isolated search lineages. Its design is guided by three principles:

\begin{itemize}
  \setlength{\itemsep}{1pt}
  \setlength{\parsep}{0pt}
  \setlength{\parskip}{0pt}
  \setlength{\topsep}{2pt}
  \item \textbf{Breadth and depth.} Breadth expands hypothesis coverage, while depth repeatedly implements, reviews, and hardens promising candidates.
  \item \textbf{Independent validation.} Held-out evidence is evaluated only after a candidate is frozen and never returns to search, preventing validation failures from being patched into task-specific solutions.
  \item \textbf{Scalable orchestration.} Isolated, disposable lineages let the funnel expand across more ideas and environments without coupling failures across candidates.
\end{itemize}

Guided by these principles, we scale AI-led auto-research across roughly $\sim$150 proposed directions and $\sim$500 executable environments, comprising more than 3,000 runs and more than 60,000 agent--environment interactions. The search yields four mechanisms that together form SoL-Pi. On EdgeBench~\cite{bytedance2026edgebench}, SoL-Pi achieves performance comparable to Pi across GPT-5.6 Sol and Opus 5 while reducing recorded token traffic by 44.7--49.0\% and API cost by about one third. Beyond these results, SoL-Pi suggests that the lasting value of RSI may lie in a search process that scales across public environments to discover reusable improvements.

\section{Method}
\label{sec:method}

\subsection{Harness Auto-Research for Token Efficiency}
\label{subsec:harness-auto-research-token-efficiency}

Our search pipeline frames harness improvement as an RSI-inspired search for reusable efficiency mechanisms, starting from a broad pool of agent-generated hypotheses. The research agent analyzes execution trajectories from the base harness to identify recurring sources of overhead, builds an idea pool of candidate harness changes, and tests their effects in development environments. To qualify for acceptance, a mechanism must improve efficiency beyond a single development environment while preserving the agent's ability to complete the required work.

Before experimentation begins, capability metrics, acceptable tolerances, and efficiency metrics are fixed and remain unchanged throughout the search. These metrics and tolerances are strictly isolated from the optimizing agent's control to prevent it from gaming the acceptance criteria. Candidate selection applies two sequential gates: every capability metric must remain within its predeclared tolerance, and the candidate must improve at least one declared efficiency metric. Among candidates that pass both gates, the pipeline retains the nondominated results under the declared metrics.
Since each mechanism is explored independently, integration remains part of the \emph{SoL-Pi: Scaling Auto-Research Loop} workflow in \cref{fig:teaser}(a): we combine retained mechanisms into SoL-Pi and refine its hyperparameters and implementation while preserving capability.

We reserve EdgeBench for final validation, keeping it isolated from the search process. The harness and acceptance rule are frozen before evaluation. Held-out results never feed back into the Auto-Research Loops: a failed validation rejects the candidate without triggering further optimization.

\subsection{Broad-to-Deep Harness Search}
\label{subsec:broad-to-deep-harness-search}

\begin{figure}[t]
  \centering
  \includegraphics[width=0.98\linewidth]{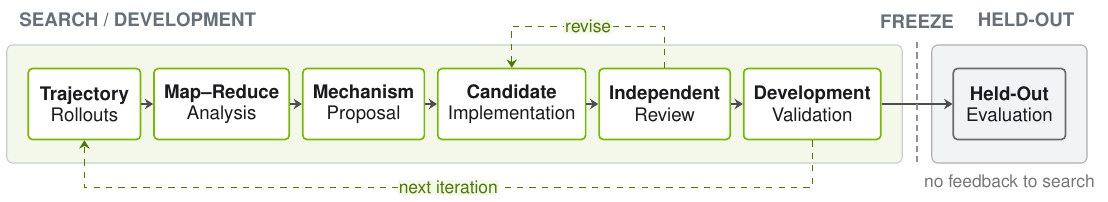}
\caption{\textbf{Development feedback and held-out evaluation remain separate throughout the search.} Execution trajectories guide mechanism proposals and implementation, while independent review and development validation inform revisions. Held-out evaluation occurs only after the candidate is frozen, and its results never feed back into the search loop.}
  \label{fig:research-lineage-loop}
\end{figure}

Our search pipeline allocates research effort in two stages: an outer stage explores a broad pool of mechanism hypotheses, and an inner stage develops selected hypotheses independently. The outer search begins with 152 proposed directions across six proposal families: context, progress, tools, delegation, prompt and policy, and improvement and evaluation. Before rollout budgets are assigned, Oracle Analysis examines existing development trajectories to identify avoidable work in the base harness. Each selected direction must identify a concrete source of overhead and propose a harness change to address it. Independent development allows unpromising directions to terminate without affecting other experiments.
The proposal families classify hypotheses by their origin rather than constrain where changes are implemented. For example, ObservationPack originates from two context hypotheses but ultimately modifies the observation boundary. This distinction helps organize the search while leaving the architecture of each mechanism open.

Each independent search follows the conventional autoresearch cycle: propose a change, implement it, run a fixed experiment, inspect the result, and retain, revise, or discard the candidate~\cite{karpathy2026autoresearch}. We extend this cycle with an iterative implementation loop based on the Ralph Loop~\cite{anthropic2026ralph}. The implementer refines the candidate to meet an explicit completion criterion, then an independent reviewer checks it before evaluation. Failed reviews trigger revision.
Each iteration may yield multiple exploration trajectories. Independent analyzers examine one trajectory each for repeated actions, context growth, large observations, or sparse diagnostic signals. A reducer combines their findings into a candidate-level summary that guides the next proposal. \Cref{fig:research-lineage-loop} summarizes the workflow and its isolation boundary.

We run each independent search as a disposable instance of a shared skill template containing a minimal research loop and operating instructions. Each experiment copies the template, sets its parameters, and runs to completion, retaining the candidate and evidence while discarding modified orchestration code. Search breadth comes from running more isolated loops; depth comes from repeated refinement within each loop. These counts describe the scope of our search; they do not establish a scaling law.

\subsection{Search Environments}
\label{subsec:search-environments}

Our search pipeline uses separate environments for mechanism discovery and transfer evaluation. The search set comprises 535 executable environments: 495 repository tasks derived from GitHub issue–pull request pairs and 40 synthetic tasks with executable success verifiers. Together, they support automatically evaluated search grounded in real software changes and open-ended problem solving.

\paragraph{Repository-derived environments.}
Each environment pairs a GitHub issue with its pre-fix repository state and offline dependencies, using the accepted patch and change history as a reference trajectory. The pull request and regression test are hidden from the agent. We retain only environments whose test fails before the patch and passes afterward.

\begin{figure}[H]
  \centering
  \includegraphics[width=0.98\linewidth]{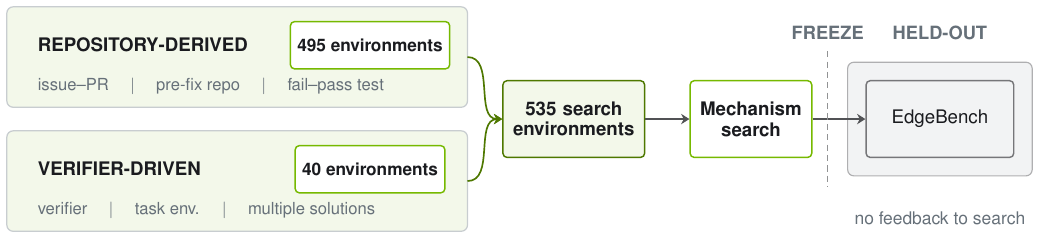}
\caption{\textbf{Two environment families support mechanism discovery while keeping EdgeBench held out.} Repository-derived environments use pre-fix repositories and hidden regression tests verified to fail before the accepted patch and pass afterward. Verifier-driven environments allow multiple solution paths under executable success criteria. EdgeBench remains isolated from search and is used only after the harness is frozen.}
  \label{fig:search-environments-heldout}
\end{figure}

\paragraph{Verifier-driven environments.}
We first generate an executable verifier that defines success, then construct a task environment around it. This allows multiple valid solution paths without requiring a reference trajectory. The 40 synthetic environments primarily use a Terminal-Bench-2-style verifier interface~\cite{merrill2026terminalbench}, extending the search beyond repository histories while preserving automatic evaluation.
\Cref{fig:search-environments-heldout} summarizes both construction paths and the isolation boundary between mechanism search and held-out evaluation.

\subsection{Discovered Harness Mechanisms}
\label{subsec:discovered-harness-mechanisms}

\begin{figure}[!t]
  \centering
  \includegraphics[width=0.98\linewidth]{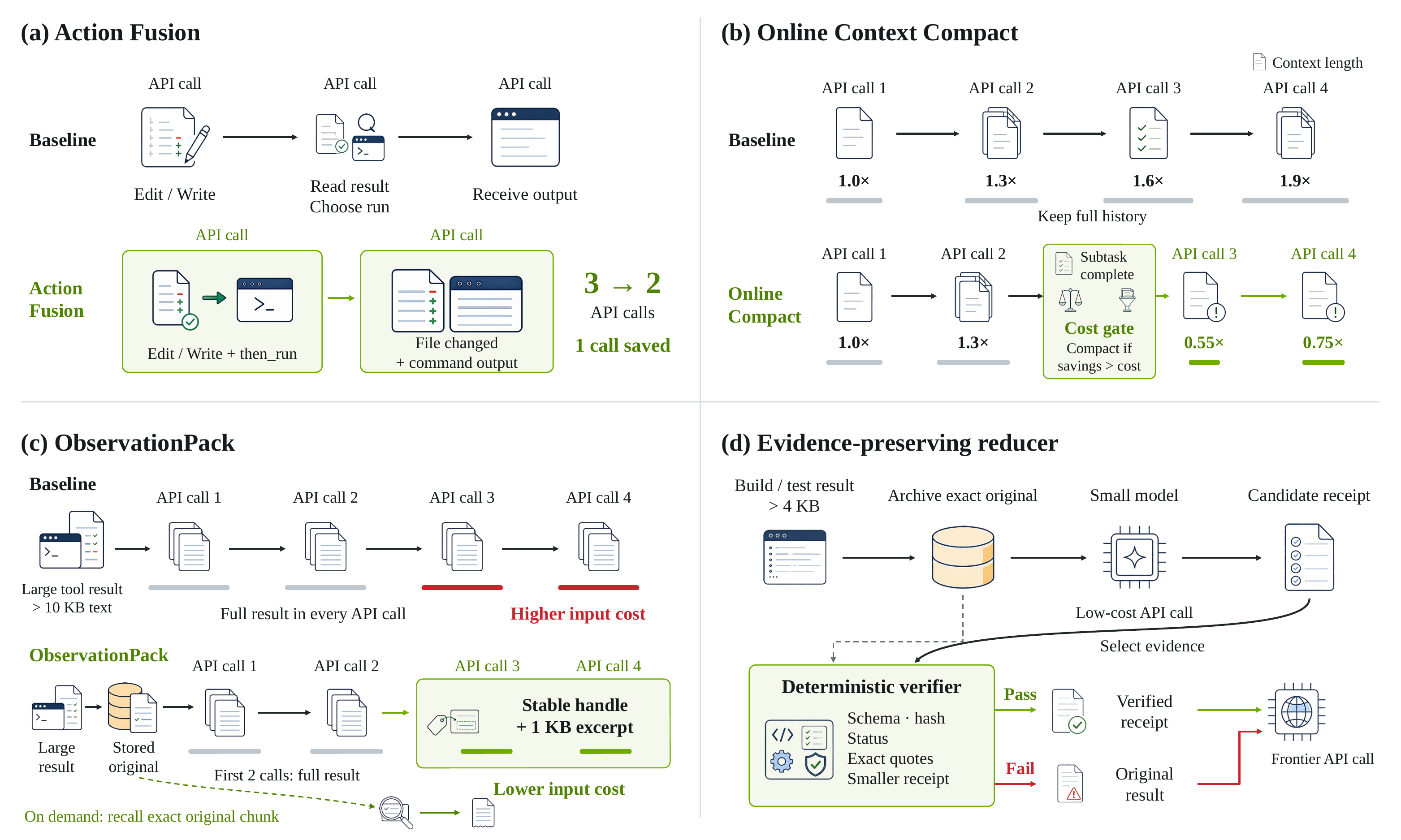}
\caption{\textbf{The four retained mechanisms act at different points in the agent--environment loop.}
(a) Action Fusion combines a mutation and its follow-up command into one request, reducing API calls from three to two.
(b) Online Context Compact evaluates compaction at subtask completion and applies it only when projected savings exceed the rewrite cost.
(c) ObservationPack sends large results in full for the first two provider requests, then replaces them with a stable handle and a 1 KB excerpt. Exact original chunks remain available on demand.
(d) Evidence-Preserving Reducer uses a low-cost model to compact results, with deterministic verification and fallback to the original on failure.
Grey marks baseline traffic, green marks mechanism paths and savings, and red marks excess cost or verification failure.}
  \label{fig:harness-outcome-map}
\end{figure}

The search yields four reusable mechanisms that pass capability-constrained selection: Action Fusion, Online Context Compact, ObservationPack, and Evidence-Preserving Reducer. They target action execution, context management, observation storage, and delegated reading, respectively, reducing repeated work while preserving information needed for later decisions. \Cref{fig:harness-outcome-map} shows their execution paths and fallback boundaries.

\paragraph{Action Fusion.}
Base Pi often edits a file and then issues a separate command to test, build, or run it. Action Fusion combines both actions into one tool request and returns their outcomes in a single observation, eliminating an intermediate model round trip. Commands that require inspecting the mutation result remain separate.

\paragraph{Online Context Compact.}
Online Context Compact uses plan-step completion to reconsider when to compact the context. The agent maintains its plan through \texttt{update\_plan}. At each completion boundary, the harness estimates remaining model requests from the observed requests between completed steps and the number of unfinished steps. It caps this estimate by the requests that would fill the current context window at the observed growth rate. The cost gate compares projected input savings with the estimated extra cost of rewriting the prompt cache. Later compactions also account for unrecovered rewrite costs and require a larger savings margin. At these boundaries, the harness invokes Pi's native compaction when this gate passes or when context usage approaches the window limit, provided compaction can shorten the context.

\paragraph{ObservationPack.}
Large tool outputs can recur in later requests even when little of their content remains relevant. ObservationPack locally archives results exceeding the threshold (10~KiB) and sends them in full for the next two provider requests. From the third request onward, it substitutes a stable handle, the original size, and a short excerpt of complete head and tail lines. The agent can retrieve exact pages through the handle as needed. Smaller results remain unchanged.

\paragraph{Evidence-Preserving Reducer.}
Evidence-Preserving Reducer compresses build and test logs of at least 4~KiB from a predefined set of commands. File reads and search results bypass the reducer. The harness archives the exact output and asks a lower-cost model to extract key evidence into a compact receipt. A deterministic verifier checks the receipt's schema, source hash, exit status, exact quotes, and size. The harness falls back to the original log if verification fails, credentials are suspected, or the receipt provides no size reduction. The reducer processes tool results before ObservationPack projects the model context; ObservationPack recognizes the reducer's receipt marker and skips those results to preserve the verified evidence. The auxiliary model extracts evidence, while the main agent retains responsibility for diagnosis and action selection.

The four mechanisms support different stages of the agent's workflow. When the agent edits code, Action Fusion combines the edit with a follow-up command. When the environment returns output, Evidence-Preserving Reducer extracts verified evidence from build and test logs, while ObservationPack avoids repeatedly sending large results in full. When the agent completes a plan step, Online Context Compact checks whether compacting the accumulated context would save tokens. These mechanisms therefore address complementary sources of overhead. We evaluate the combined harness end to end to verify their joint effect on capability and efficiency.

\subsection{Implementation Details and Backend Setup}
\label{subsec:implementation-details}

We implement all four mechanisms as extensions to Pi. Action Fusion adds an optional follow-up command to file-mutation tools and returns both outcomes in one observation. Online Context Compact checks the cache-cost gate at plan-step completion and also supports compaction near the context limit. The gate estimates cache-rewrite overhead from the context size and the cache-write/read price ratio; it does not separately price the summarization call. ObservationPack archives large outputs locally and provides stable handles for exact retrieval. Evidence-Preserving Reducer uses GPT-5.6 Luna at \texttt{high} to extract evidence and verifies the resulting receipt before passing it to the main agent.

Before held-out evaluation, we freeze the mechanism source, configuration, metrics, capability tolerances, and acceptance rule. All EdgeBench outputs remain outside the search loop. Of EdgeBench's 51 public tasks, 11 are used for one-way acceptance of frozen candidates; the remaining 40 are reserved for final evaluation of generalization.

\section{Experimental Evaluation}
\label{sec:evaluation}
\label{sec:experimental-evaluation}

We evaluate \solpi on \edgebench~\cite{bytedance2026edgebench}, Terminal-Bench 4~\cite{marten2026terminalbench4}, IMO 2026~\cite{imo2026problems}, and a kernel-optimization benchmark~\cite{anthropicOriginalPerformanceTakehome}. We report average score, token traffic (in billions), and API cost. Token efficiency is measured as API cost per unit of aggregate task score.
\Cref{sec:comparison} compares \solpi with native and third-party harnesses and tests whether it transfers from GPT-5.6 Sol to Opus 5 without modification. \Cref{sec:terminal-bench-4} reports Terminal-Bench 4 completion rates and IMO 2026 Lean 4-verified problem counts, together with costs for both. \Cref{sec:agent-swarm} evaluates \solpi in a coordinated agent swarm. \Cref{sec:learned-mechanisms-generalization} analyzes individual and combined mechanism effects and activation patterns across backends and configurations. \Cref{sec:learning-process} traces the development and selection of \actionfusion.

\subsection{Overall Comparison}
\label{sec:comparison}
\enlargethispage{\baselineskip}

\edgebench currently releases 51 of its 134 tasks as open source, and our evaluation uses this public set~\cite{bytedance2026edgebench}. \Cref{tab:main-comparison} compares eight evaluated configurations under GPT-5.6 Sol and lists each primary backend; the EdgeBench official GPT-5.5 row is an unranked score-only reference. We report two SoL-Pi operating points. \solpi [Efficiency] is the fixed complete four-mechanism stack, reported as the token-efficiency-oriented point. \solpi [Performance] is the single-mechanism configuration with the highest average score, selected separately for each backend from \cref{tab:component-ablation}; under GPT-5.6 Sol, it corresponds to \observationpack.

The Efficiency point uses a total of 1.10~B tokens, 49.0\% fewer than Pi, while retaining 93.7\% of Pi's average score (42.0 vs.\ 44.8). Its token cost is 33.2\% lower than Pi's. The Performance point raises average score from 44.8 to 47.2, a 5.3\% gain, while reducing token traffic by 6.1\% and improving token efficiency by 9.8\%.

\begin{table}[H]
    \centering
    \normalfont\fontsize{7.0pt}{8.4pt}\selectfont
    \setlength{\tabcolsep}{1.35pt}
    \renewcommand{\arraystretch}{1.18}
    \caption{\textbf{Harness comparison on \edgebench.} \solpi [Efficiency] is the fixed complete-stack point selected for token efficiency, while \solpi [Performance] is the GPT-5.6 Sol single-mechanism configuration with the highest average score in \cref{tab:component-ablation}. Best and next-distinct values among the eight measured rows are bold and underlined. The EdgeBench official GPT-5.5 @2h checkpoint (31.2) is unranked~\cite{bytedance2026edgebenchleaderboard}; dashes denote unreported token traffic and token cost.}
    \label{tab:main-comparison}
    \begin{tabularx}{\linewidth}{@{}>{\raggedright\arraybackslash}p{0.205\linewidth}>{\raggedright\arraybackslash}p{0.14\linewidth}*{8}{>{\centering\arraybackslash}X}@{}}
        \toprule
        \multirow{2}{*}{\textbf{Harness}}
        & \multirow{2}{*}{\textbf{Backend}}
        & \multicolumn{5}{c}{\textbf{Recorded Token Traffic (B)}}
        & \multirow{2}{*}{\shortstack{\textbf{Token Cost}\\\textbf{(\$)}$\downarrow$}}
        & \multirow{2}{*}{\shortstack{\textbf{Avg.}\\\textbf{Score}$\uparrow$}}
        & \multirow{2}{*}{\shortstack{\textbf{Token Eff.}\\\textbf{(\$/score)}$\downarrow$}} \\
        \cmidrule(lr{0.25em}){3-7}
        & & \textbf{Input}$\downarrow$ & \textbf{Cache R.}$\downarrow$ & \textbf{Cache W.}$\downarrow$ & \textbf{Output}$\downarrow$ & \textbf{Total}$\downarrow$ & & & \\
        \midrule
        EdgeBench official @2h & GPT-5.5 & -- & -- & -- & -- & -- & -- & 31.2 & -- \\
        \midrule
        Codex~\cite{openai2026codexcli} & GPT-5.6 Sol & 0.0053 & 3.0287 & \underline{0.0145} & \underline{0.0052} & 3.0537 & 1{,}787 & 34.738 & 1.0086 \\
        OpenSquilla~\cite{tokenrhythm2026opensquilla} & GPT-5.6 Sol & \textbf{0.0001} & \underline{1.2533} & 0.0776 & \textbf{0.0044} & \underline{1.3353} & \underline{1{,}243} & 24.506 & 0.9945 \\
        Oh-My-Pi~\cite{can13572026ohmypi} & GPT-5.6 Sol & 0.1135 & 2.0448 & 0.0484 & 0.0168 & 2.2235 & 1{,}832 & 26.921 & 1.3347 \\
        OpenCode~\cite{anomaly2026opencode} & GPT-5.6 Sol & 0.0012 & 2.2625 & 0.2865 & 0.0164 & 2.5668 & 3{,}422 & 29.552 & 2.2704 \\
        Oh-My-Opencode~\cite{codeyeongyu2026ohmyopencode} & GPT-5.6 Sol & 0.0044 & 2.4373 & 0.1286 & 0.0121 & 2.5825 & 2{,}678 & 38.523 & 1.3633 \\
        Pi~\cite{earendil2026pi} & GPT-5.6 Sol & 0.0011 & 2.1326 & \textbf{0.0141} & 0.0059 & 2.1538 & 1{,}339 & \underline{44.833} & 0.5855 \\
        \midrule
        \rowcolor{nvgreen}
        \textbf{\solpi [Efficiency]} & GPT-5.6 Sol & 0.0009 & \textbf{1.0605} & 0.0316 & 0.0061 & \textbf{1.0990} & \textbf{894} & 42.003 & \textbf{0.4174} \\
        \rowcolor{nvgreenlight}
        \textbf{\solpi [Performance]} & GPT-5.6 Sol & \underline{0.0002} & 2.0005 & 0.0160 & 0.0056 & 2.0224 & 1{,}271 & \textbf{47.208} & \underline{0.5280} \\
        \bottomrule
    \end{tabularx}
    \vspace{1pt}
    \parbox{\linewidth}{\raggedright\fontsize{6.5pt}{7.5pt}\selectfont\textit{Note.} API prices change over time; all token costs in this paper use the API prices as of August 17, 2026.}
\end{table}

\begin{table}[!t]
    \centering
    \normalfont\fontsize{7.2pt}{8.7pt}\selectfont
    \setlength{\tabcolsep}{2pt}
    \renewcommand{\arraystretch}{1.24}
\caption{\textbf{SoL-Pi configurations and transfer across models on \edgebench.} SoL-Pi [Efficiency] combines all four mechanisms and transfers from GPT-5.6 Sol to Opus 5 without further search or adaptation. SoL-Pi [Performance] uses the highest-scoring single mechanism for each model in \cref{tab:component-ablation}: ObservationPack for GPT-5.6 Sol and Action Fusion for Opus 5. Within each model block, best values are \textbf{bold} and next-distinct values are \underline{underlined}.}
    \label{tab:backend-transfer}
    \begin{tabularx}{\linewidth}{@{}>{\raggedright\arraybackslash}p{0.22\linewidth}*{8}{>{\centering\arraybackslash}X}@{}}
        \toprule
        \multirow{2}{*}{\textbf{Harness}}
        & \multicolumn{5}{c}{\textbf{Recorded Token Traffic (B)}}
        & \multirow{2}{*}{\shortstack{\textbf{Token Cost}\\\textbf{(\$)}$\downarrow$}}
        & \multirow{2}{*}{\shortstack{\textbf{Avg.}\\\textbf{Score}$\uparrow$}}
        & \multirow{2}{*}{\shortstack{\textbf{Token Eff.}\\\textbf{(\$/score)}$\downarrow$}} \\
        \cmidrule(lr{0.25em}){2-6}
        & \textbf{Input}$\downarrow$ & \textbf{Cache R.}$\downarrow$ & \textbf{Cache W.}$\downarrow$ & \textbf{Output}$\downarrow$ & \textbf{Total}$\downarrow$ & & & \\
        \midrule
        \multicolumn{9}{c}{\textit{\textbf{GPT-5.6 Sol (Search Backend)}}} \\
        \midrule
        Codex~\cite{openai2026codexcli} & 0.0053 & 3.0287 & \underline{0.0145} & \textbf{0.0052} & 3.0537 & 1{,}787 & 34.738 & 1.0086 \\
        Pi~\cite{earendil2026pi} & 0.0011 & 2.1326 & \textbf{0.0141} & 0.0059 & 2.1538 & 1{,}339 & \underline{44.833} & 0.5855 \\
        \midrule
        \rowcolor{nvgreen}
        \textbf{\solpi [Efficiency]} & \underline{0.0009} & \textbf{1.0605} & 0.0316 & 0.0061 & \textbf{1.0990} & \textbf{894} & 42.003 & \textbf{0.4174} \\
        \rowcolor{nvgreenlight}
        \textbf{\solpi [Performance]} & \textbf{0.0002} & \underline{2.0005} & 0.0160 & \underline{0.0056} & \underline{2.0224} & \underline{1{,}271} & \textbf{47.208} & \underline{0.5280} \\
        \midrule
        \multicolumn{9}{c}{\textit{\textbf{Opus 5 (Held-Out Backend)}}} \\
        \midrule
        Claude Code~\cite{anthropic2026claudecode} & \underline{0.0002} & \underline{1.8116} & 0.1701 & 0.0226 & \underline{2.0045} & 2{,}535 & 43.689 & 1.1377 \\
        Pi~\cite{earendil2026pi} & \textbf{0.0000} & 2.3203 & \textbf{0.0348} & 0.0145 & 2.3697 & 1{,}741 & \underline{44.756} & 0.7625 \\
        \midrule
        \rowcolor{nvgreen}
        \textbf{\solpi [Efficiency]} & \textbf{0.0000} & \textbf{1.2626} & \underline{0.0352} & \textbf{0.0123} & \textbf{1.3101} & \textbf{1{,}158} & 42.224 & \textbf{0.5376} \\
        \rowcolor{nvgreenlight}
        \textbf{\solpi [Performance]} & \textbf{0.0000} & 2.0520 & \underline{0.0352} & \underline{0.0144} & 2.1016 & \underline{1{,}605} & \textbf{50.482} & \underline{0.6235} \\
        \bottomrule
    \end{tabularx}
\end{table}

\Cref{tab:backend-transfer} compares the native harness, Pi, \solpi [Efficiency], and \solpi [Performance] on GPT-5.6 Sol and Opus 5. To test transfer, we apply \solpi, developed with GPT-5.6 Sol, to Opus 5 without further search or adaptation. On Opus 5, it retains 94.3\% of Pi's average score while reducing token traffic by 44.7\% and API cost by 33.5\%, relative to Pi's point estimates. These results, alongside the activation patterns in \cref{sec:learned-mechanisms-generalization}, provide preliminary evidence of transfer to an unseen LLM backend.

\Cref{fig:teaser}(b) summarizes the native-harness, Pi, and complete-stack SoL-Pi results on EdgeBench across the two backends. These bars are example results from our broader benchmark evaluation; Terminal-Bench 4 and IMO 2026 are reported in \cref{sec:terminal-bench-4}.

\subsection{Evaluation on Terminal-Bench 4 and IMO 2026}
\label{sec:terminal-bench-4}

We also compare Codex, Pi, and \solpi on 63 CPU-only tasks from Terminal-Bench 4~\cite{marten2026terminalbench4}. \Cref{tab:terminal-bench-4} reports the number of solved tasks, total model cost, and cost per solved task, with all costs reported as API costs. Codex and Pi each solve 18 tasks, while \solpi solves 15. Compared with Pi, \solpi reduces total model cost by 26.3\% (\$211.12 vs.\ \$286.45) and cost per solved task by 11.6\% (\$14.07 vs.\ \$15.91). These results suggest that the harness's efficiency gains generalize beyond EdgeBench, with lower total cost.

We further evaluate the three harnesses on IMO 2026~\cite{imo2026problems} using GPT-5.6 Sol (\texttt{xhigh}), requiring each solution to be formalized and verified in Lean 4. \solpi passes three of six problems at a total model cost of \$62.69, achieving the lowest cost per passed problem (\$20.90), compared with \$22.89 for Codex and \$25.32 for Pi.

\begin{table}[!t]
    \centering
    \normalfont\fontsize{7pt}{8.5pt}\selectfont
    \setlength{\tabcolsep}{1.5pt}
    \renewcommand{\arraystretch}{1.2}
    \caption{\textbf{Harness comparison on Terminal-Bench 4 and IMO 2026.} Terminal-Bench 4 uses 63 CPU-only tasks; the IMO evaluation covers six problems. All IMO runs use GPT-5.6 Sol (\texttt{xhigh}). Costs are in USD, and best values within each benchmark are \textbf{bold}.}
    \label{tab:terminal-bench-4}
    \begin{tabularx}{\linewidth}{@{}l*{6}{>{\centering\arraybackslash}X}@{}}
        \toprule
        \multirow{2}{*}{\textbf{Harness}}
        & \multicolumn{3}{c}{\textbf{Terminal-Bench 4}}
        & \multicolumn{3}{c}{\textbf{IMO 2026}} \\
        \cmidrule(lr){2-4}\cmidrule(lr){5-7}
        & \shortstack{\textbf{Solved Tasks}\\\textbf{(out of 63)}$\uparrow$}
        & \shortstack{\textbf{Total Model}\\\textbf{Cost (\$)}$\downarrow$}
        & \shortstack{\textbf{Cost / Solved}\\\textbf{Task (\$)}$\downarrow$}
        & \shortstack{\textbf{Pass}\\\textbf{(out of 6)}$\uparrow$}
        & \shortstack{\textbf{Total Model}\\\textbf{Cost (\$)}$\downarrow$}
        & \shortstack{\textbf{Cost / Passed}\\\textbf{Problem (\$)}$\downarrow$} \\
        \midrule
        Codex~\cite{openai2026codexcli} & \textbf{18} & 272.35 & 15.13 & \textbf{5} & 114.47 & 22.89 \\
        Pi~\cite{earendil2026pi} & \textbf{18} & 286.45 & 15.91 & 3 & 75.95 & 25.32 \\
        \rowcolor{nvgreen}
        \textbf{\solpi} & 15 & \textbf{211.12} & \textbf{14.07} & 3 & \textbf{62.69} & \textbf{20.90} \\
        \bottomrule
    \end{tabularx}
    \vspace{2pt}
    \parbox{\linewidth}{\raggedright\fontsize{7pt}{8.5pt}\selectfont\textit{Note.} GPU-dependent Terminal-Bench 4 tasks were excluded due to infrastructure limits. The IMO evaluation uses AxiomMath/IMO2026's formal statements and follows Humanfia's verification setup~\cite{humanfia2026hoa,axiommath2026imo}. Each problem is capped at 150 minutes to limit unproductive looping, and costs include all model activity within this budget.}
\end{table}

\subsection{Efficient Agent Swarms}
\label{sec:agent-swarm}

We evaluate \solpi in a multi-agent kernel-optimization experiment measured in simulated machine cycles~\cite{anthropicOriginalPerformanceTakehome}. We compare three configurations in one two-hour run each: a single Codex agent, a Codex coordinator with 20 Pi baseline workers, and a Codex coordinator with 20 \solpi workers using all four mechanisms. The single agent and coordinators use GPT-5.6 Sol at \texttt{xhigh}; all workers use GPT-5.6 Luna at \texttt{xhigh}. Every run starts from the same frozen starter, which requires 147{,}734 cycles, with fresh sessions and no solutions or notes from earlier runs.

In both swarms, workers form five groups of four, with independent workspaces and a shared evidence board within each group. Workers exchange notes to reproduce or combine promising findings, while the coordinator relays findings across groups. The shared best result is updated only when the coordinator submits an immutable candidate snapshot and independent verification confirms a strict improvement.

\begin{figure}[H]
    \centering
    \includegraphics[width=\linewidth]{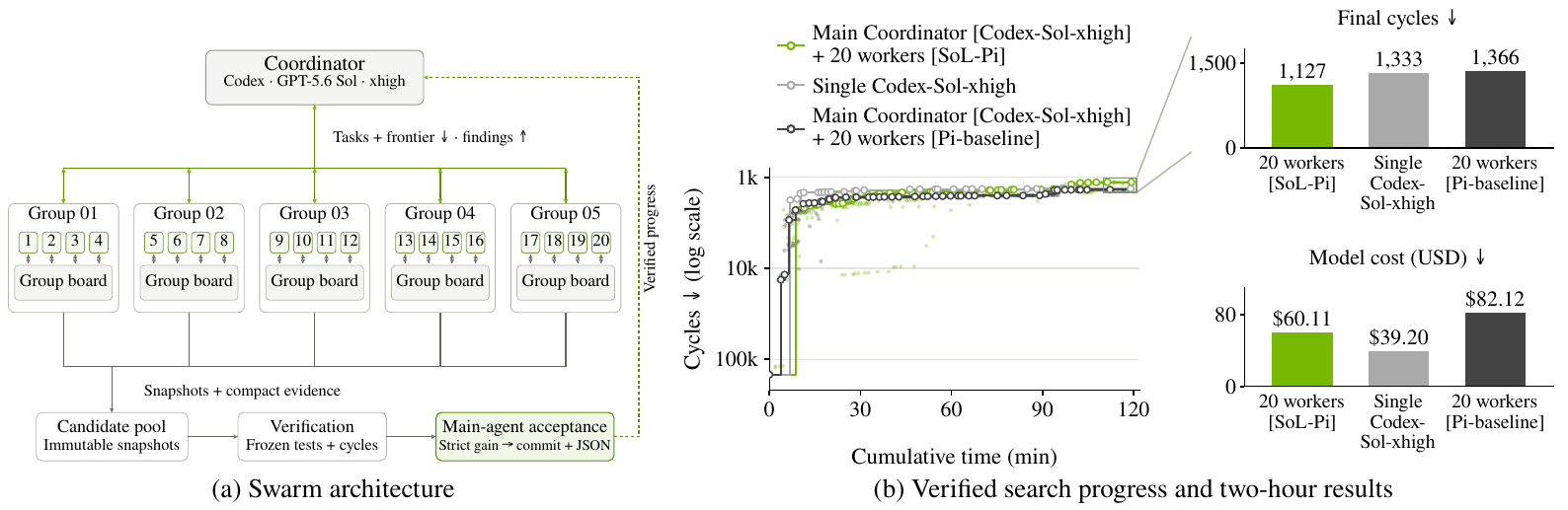}
    \caption{\textbf{Agent swarm architecture and two-hour search results.}
    (a) Group boards support local collaboration, and the coordinator shares findings across groups and submits candidates for independent verification.
    (b) Filled points show correct submitted candidates, open circles mark accepted commits, and step lines track the best accepted result. The cost chart shows cumulative API cost.}
    \label{fig:agent-swarm}
\end{figure}

\Cref{fig:agent-swarm}(b) shows that the \solpi swarm reaches 1{,}127 cycles at \$60.11, compared with 1{,}333 cycles at \$39.20 for the single agent and 1{,}366 cycles at \$82.12 for the Pi baseline swarm. Under the same execution budget, the \solpi swarm reduces API cost by 26.8\% relative to the Pi baseline swarm, while the single agent remains least expensive. All final candidates pass the official correctness check. The \solpi swarm and single agent pass all eight speed thresholds, whereas the Pi baseline swarm passes seven, missing the final threshold of fewer than 1{,}363 cycles.
These results suggest that the value of \solpi may extend beyond individual agents: a more efficient harness could help agent swarms and multi-agent systems turn a fixed budget into more effective collective exploration.

\subsection{Learned Mechanisms and Backend Behavior}
\label{sec:learned-mechanisms-generalization}

We assess the standalone contribution of each independently learned mechanism through an add-one evaluation. \Cref{tab:component-ablation} compares the Pi baseline, four variants that each add a single mechanism to Pi, and the complete \solpi stack under both model backends.
Every component reduces the total token count under both backends. \observationpack records the highest average score under GPT-5.6 Sol, while \actionfusion does so under Opus 5. The complete stack represents the efficiency-oriented point and has the lowest total token count and token cost in both backend blocks.

\begin{figure}[H]
    \centering
    \includegraphics[width=0.96\linewidth]{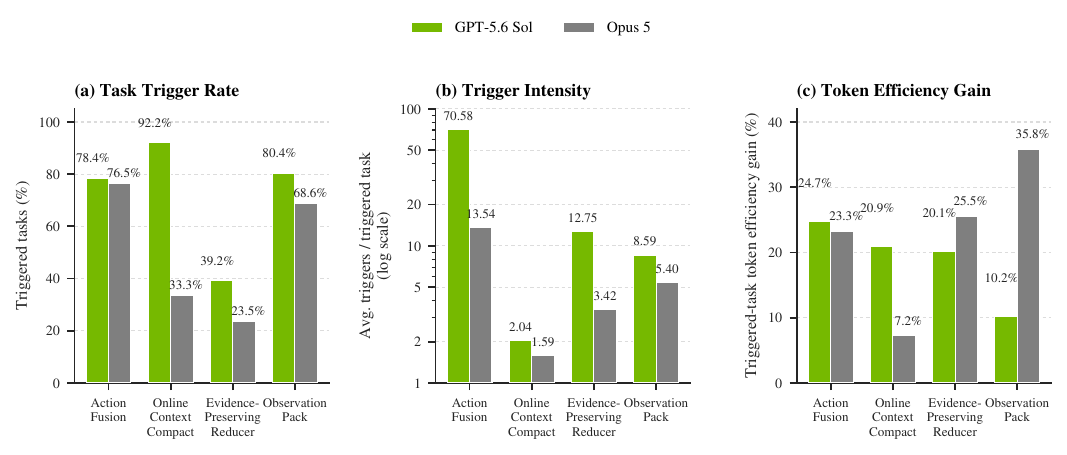}
    \caption{\textbf{Backend-dependent activation of the retained mechanisms on \edgebench.} Panels show trigger rate, trigger intensity on a logarithmic scale, and token-efficiency gain.}
    \label{fig:component-backend-activation}
\end{figure}

\Cref{fig:component-backend-activation} shows that mechanism activation varies across backends. Trigger rate is the fraction of tasks on which a mechanism activates, while trigger intensity is the mean number of activations per triggered task. Both are lower under Opus 5, which may reflect the harness being optimized exclusively on GPT-5.6 Sol trajectories. Nevertheless, every configuration evaluated under Opus 5 improves token efficiency on its triggered tasks, and the complete stack preserves a similar aggregate score--efficiency trade-off across the two backends (\cref{tab:backend-transfer}).

\begin{figure}[!t]
    \centering
    \includegraphics[width=\linewidth]{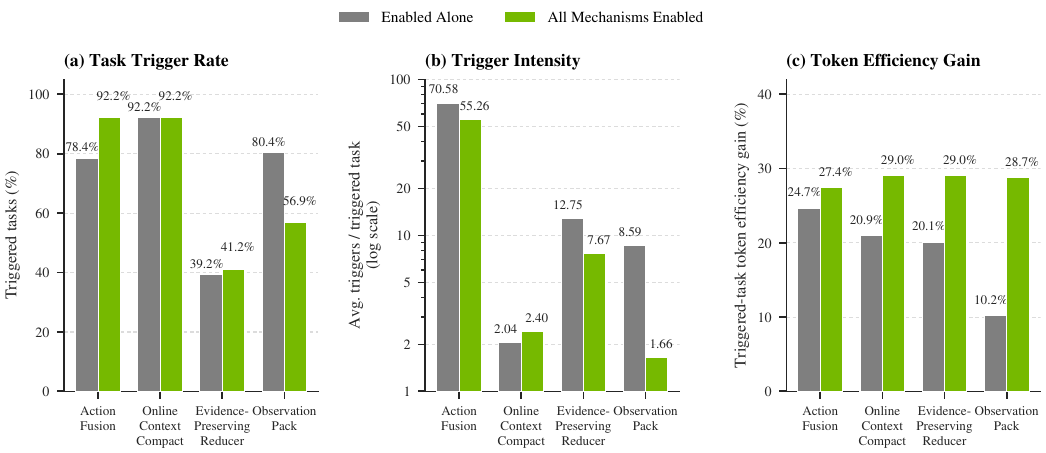}
    \caption{\textbf{Merge behavior of independently explored mechanisms under GPT-5.6 Sol (\texttt{xhigh}).} Paired bars compare standalone and full-stack trigger rate, trigger intensity, and token-efficiency gain for each mechanism. Gains use each configuration's own triggered-task subset and corresponding disabled baseline, so comparisons are descriptive. \observationpack becomes more selective in the full stack, while every mechanism shows a larger token-efficiency gain than in its standalone configuration, a pattern consistent with complementarity in the evaluated stack.}
    \label{fig:component-configuration-comparison}
\end{figure}

To assess the effects of combining mechanisms, \cref{fig:component-configuration-comparison} compares each standalone configuration with the full stack. \observationpack becomes more selective in the full stack, potentially due to overlap with \lunaepdr on observation-heavy trajectories. For every mechanism, the full stack achieves a larger token-efficiency gain than the standalone configuration on their respective triggered-task subsets. This pattern is consistent with complementarity in the evaluated stack, although comparisons on each configuration's own triggered-task subset do not isolate interaction effects.

\begin{table}[H]
    \centering
    \normalfont\fontsize{7.2pt}{8.7pt}\selectfont
    \setlength{\tabcolsep}{2pt}
    \renewcommand{\arraystretch}{1.24}
\caption{\textbf{Single-component evaluation on \edgebench.} Each component row adds one mechanism to Pi; \solpi [Efficiency] combines all four. Light green and $\dagger$ identify the configuration selected as \solpi [Performance] for each backend; dark green highlights \solpi [Efficiency]. Token cost and token efficiency are computed using fixed API prices. Best values are \textbf{bold}, and the next-best distinct values are \underline{underlined}.}
    \label{tab:component-ablation}
    \begin{tabularx}{\linewidth}{@{}>{\raggedright\arraybackslash}p{0.22\linewidth}*{8}{>{\centering\arraybackslash}X}@{}}
        \toprule
        \multirow{2}{*}{\textbf{Configuration}}
        & \multicolumn{5}{c}{\textbf{Recorded Token Traffic (B)}}
        & \multirow{2}{*}{\shortstack{\textbf{Token Cost}\\\textbf{(\$)}$\downarrow$}}
        & \multirow{2}{*}{\shortstack{\textbf{Avg.}\\\textbf{Score}$\uparrow$}}
        & \multirow{2}{*}{\shortstack{\textbf{Token Eff.}\\\textbf{(\$/score)}$\downarrow$}} \\
        \cmidrule(lr{0.25em}){2-6}
        & \textbf{Input}$\downarrow$ & \textbf{Cache R.}$\downarrow$ & \textbf{Cache W.}$\downarrow$ & \textbf{Output}$\downarrow$ & \textbf{Total}$\downarrow$ & & & \\
        \midrule
        \multicolumn{9}{c}{\textit{\textbf{GPT-5.6 Sol}}} \\
        \midrule
        Pi Baseline~\cite{earendil2026pi} & 0.0011 & 2.1326 & \textbf{0.0141} & 0.0059 & 2.1538 & 1{,}339 & 44.833 & 0.5855 \\
        + \actionfusion & 0.0011 & 1.8718 & 0.0180 & 0.0060 & 1.8968 & 1{,}235 & \underline{46.664} & 0.5190 \\
        + \onlinecompact & \underline{0.0008} & \underline{1.2602} & 0.0217 & \textbf{0.0055} & \underline{1.2881} & \underline{935} & 41.993 & \underline{0.4365} \\
        + \lunaepdr & 0.0051 & 1.9089 & 0.0181 & \textbf{0.0055} & 1.9375 & 1{,}200 & 44.630 & 0.5274 \\
        \rowcolor{nvgreenlight}
        + \observationpack$^{\dagger}$ & \textbf{0.0002} & 2.0005 & \underline{0.0160} & \underline{0.0056} & 2.0224 & 1{,}271 & \textbf{47.208} & 0.5280 \\
        \midrule
        \rowcolor{nvgreen}
        \textbf{\solpi [Efficiency]} & 0.0009 & \textbf{1.0605} & 0.0316 & 0.0061 & \textbf{1.0990} & \textbf{894} & 42.003 & \textbf{0.4174} \\
        \midrule
        \multicolumn{9}{c}{\textit{\textbf{Opus 5}}} \\
        \midrule
        Pi Baseline~\cite{earendil2026pi} & \textbf{0.0000} & 2.3203 & \underline{0.0348} & 0.0145 & 2.3697 & 1{,}741 & 44.756 & 0.7625 \\
        \rowcolor{nvgreenlight}
        + \actionfusion$^{\dagger}$ & \textbf{0.0000} & 2.0520 & 0.0352 & 0.0144 & 2.1016 & 1{,}605 & \textbf{50.482} & 0.6235 \\
        + \onlinecompact & \textbf{0.0000} & 1.8618 & 0.0388 & 0.0145 & 1.9152 & 1{,}537 & \underline{49.155} & 0.6130 \\
        + \lunaepdr & \textbf{0.0000} & 1.8685 & \textbf{0.0315} & 0.0131 & 1.9131 & 1{,}456 & 43.405 & 0.6578 \\
        + \observationpack & \textbf{0.0000} & \underline{1.3960} & 0.0356 & \textbf{0.0102} & \underline{1.4418} & \underline{1{,}176} & 47.047 & \textbf{0.4899} \\
        \midrule
        \rowcolor{nvgreen}
        \textbf{\solpi [Efficiency]} & \textbf{0.0000} & \textbf{1.2626} & 0.0352 & \underline{0.0123} & \textbf{1.3101} & \textbf{1{,}158} & 42.224 & \underline{0.5376} \\
        \bottomrule
    \end{tabularx}
\end{table}

\paragraph{Cache Reuse and Total Cost.}
Shortening context can reduce prompt-cache reuse when it changes a previously cached prefix~\cite{anthropic2026promptcaching}. Cached input also incurs a cost, so preserving a long prefix is not always the cheapest choice over an entire task. Online Context Compact and ObservationPack trade some prefix reuse for less repeated input. \Cref{tab:component-ablation} shows this trade-off: with GPT-5.6 Sol, the complete stack reduces cache-read traffic from 2.1326~B to 1.0605~B tokens, while cache-write traffic increases from 0.0141~B to 0.0316~B. Despite the additional cache-write traffic, total model cost falls from \$1{,}339 to \$894. Every single-mechanism configuration also lowers cost per score point relative to Pi under both backends. These results support evaluating the full task cost alongside task quality, rather than cache reuse alone.

\subsection{From Observation to a Retained Mechanism}
\label{sec:learning-process}

\begin{figure}[!t]
    \centering
    \includegraphics[width=\linewidth]{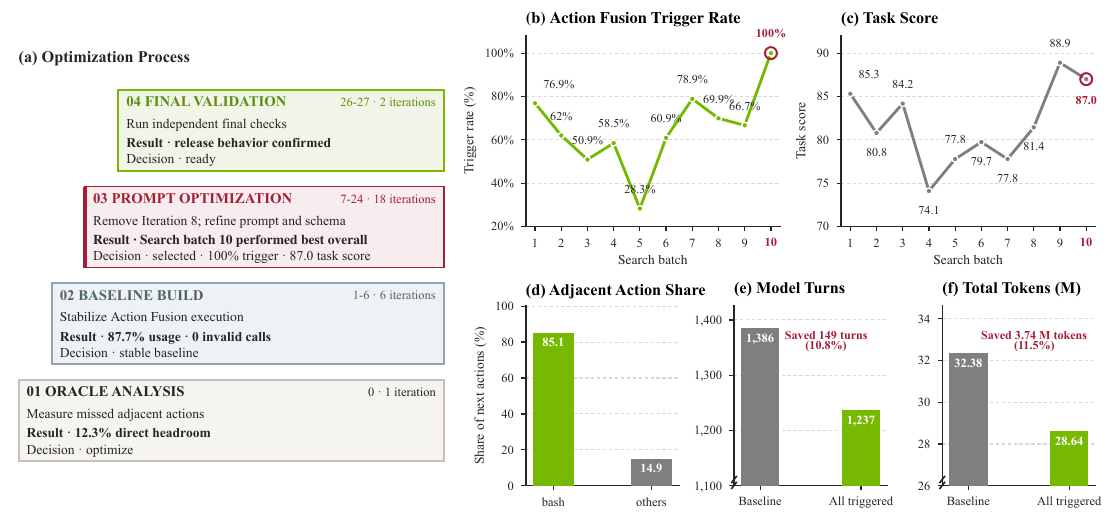}
    \caption{\textbf{\actionfusion from opportunity to retention.} Panel (a) summarizes the 27 recorded iterations across four stages. Stage 03 records 18 prompt-optimization explorations, whereas panels (b)--(c) show ten retained steps: some steps evaluate multiple candidates in parallel, and only the best result from each parallel batch is retained. Panel (d) reports observed adjacent-action composition; panels (e)--(f) project model turns and tokens under full triggering.}
    \label{fig:action-fusion-discovery}
\end{figure}

We use \actionfusion as a case study of how a candidate was discovered and retained. As shown in \cref{fig:action-fusion-discovery}, the recorded lineage spans 27 iterations across four stages: oracle analysis, baseline construction, prompt and tool-schema optimization, and final held-out validation.

The oracle-analysis panels~(d)--(f) identify repeated adjacent actions and project an 11.5\% token reduction under full triggering, motivating a dedicated \actionfusion lineage. During baseline construction, prompt-only triggering proved unreliable. The agent therefore extended the tool schema to expose the fused action directly, establishing a stable baseline with no invalid calls. It then refined the prompt and schema on development tasks, introducing trigger rate alongside task score as an intermediate acceptance metric. The final configuration was selected using both metrics, frozen, and retained after held-out validation. This case illustrates how a lineage-local auto-research loop can stabilize a mechanism's interface and triggering behavior. The agent's introduction of trigger rate also demonstrates that auto-research can develop mechanism-specific intermediate metrics to guide optimization alongside end-task performance.

\ifdefined\SoLPiRelatedWorkInAppendix
\else
  
\section{Related Work}
\label{sec:related-work}

\subsection{Agent Harnesses and Automated Agent Design}
An agent's behavior depends not only on its underlying model, but also on the harness that presents state, exposes actions, and processes feedback. SWE-agent showed that changing the agent--computer interface around a fixed model can materially affect repository-level software-engineering performance~\cite{yang2024sweagent}. Subsequent work has examined interface design, context management, tool use, and delegation as system-level choices that shape agent behavior~\cite{ning2026codeharness,lumer2026recursive}. These design choices become increasingly consequential in long-horizon tasks, where accumulated interaction histories and environment observations can increase context size and execution overhead~\cite{bytedance2026edgebench,anthropic2025longrunningharness}.

Automated agent design explores several optimization targets. GEPA improves prompts through reflection on execution trajectories~\cite{agrawal2026gepa}, while ADAS formulates agent design as code search~\cite{hu2025automated}. AFlow and AgentSquare search over workflow structures and modular agent components, respectively~\cite{zhang2025aflow,shang2025agentsquare}. Related research on recursive self-improvement has theoretical roots in the G\"odel Machine~\cite{schmidhuber2007godel}. Recent systems explore executable forms of self-modification: the Darwin G\"odel Machine iteratively modifies and evaluates agent code~\cite{zhang2026dgm}, while Hyperagents also make the meta-level improvement procedure itself editable~\cite{zhang2026hyperagents}. Our approach draws on this line of research to search for reusable efficiency improvements in harness mechanisms while keeping the underlying model fixed.

\subsection{Automated Harness Optimization}

Recent methods optimize harness code and configurations using execution feedback. AutoHarness synthesizes environment-specific code guards and, in some settings, complete code policies~\cite{lou2026autoharness}. Recursive Harness Self-Improvement (RHI) refines prompt-level specifications of the agent loop for individual tasks~\cite{lee2026rhi}. Meta-Harness searches over executable harness programs using prior code, scores, and execution traces, maintaining a Pareto frontier over task performance and context cost~\cite{lee2026metaharness}. AHE evolves modular coding-harness components from trajectory evidence and evaluates their transfer across tasks and models~\cite{lin2026ahe}. MemoHarness stores execution experience and uses it to adapt harness configurations to individual test cases at inference time~\cite{huang2026memoharness}.

The extent to which the improvements discovered generalize remains an important evaluation question. Wang et al. report limited gains on held-out tasks for the methods they evaluate and highlight the risk of overstating improvements when search and evaluation tasks overlap~\cite{wang2026rethinkingharnessevolution}. Our search pipeline explores candidates across diverse executable development environments, using a broad-to-deep funnel to refine promising changes while keeping final held-out evaluation separate from candidate development. The search targets token efficiency while checking that candidates maintain task performance. SoL-Pi combines four mechanisms selected through this process. Its held-out evaluation provides preliminary evidence that RSI-inspired search at the harness layer can discover mechanisms that generalize beyond their development environments.

\subsection{Context and Token-Efficient Agents}

Long-horizon agents can incur substantial token overhead as interaction histories and environment observations accumulate. LLM-as-Code limits context accumulation by moving deterministic control flow into executable code~\cite{qi2026llmascode}. Other methods reduce the information retained during execution: AgentDiet removes redundant and outdated trajectory content~\cite{xiao2026reducing}, while ACON optimizes the compression of observations and interaction histories~\cite{kang2025acon}. Context-Folding and AgentFold enable agents to manage their working context through trajectory folding and compression~\cite{sun2025contextfolding,ye2025agentfold}, and Context as a Tool exposes context maintenance as an explicit agent action~\cite{liu2026contexttool}. Agentic Context Engineering (ACE) focuses on accumulating and refining reusable strategies in context through execution feedback~\cite{zhang2026ace}. These approaches provide mechanisms for controlling context growth and organizing execution. Our approach studies automated discovery and selection across multiple harness components, seeking reusable changes that reduce token usage while maintaining task performance. SoL-Pi combines the selected mechanisms across action execution, context compaction, observation handling, and delegated reading.

\fi
\ifdefined\SoLPiBeforeConclusion\SoLPiBeforeConclusion\fi
\section{Conclusion}

We introduce \solpi, an RSI-inspired auto-research system for discovering and combining efficient agent-harness mechanisms. On \edgebench, the best-performing candidates improve model performance by 5.3--12.8\% and token efficiency by 9.8--18.2\%, while the complete stack reduces token traffic by 44.7--49.0\% and token cost by about one third at comparable performance. The complete stack's consistent results across GPT-5.6 Sol and Opus 5 demonstrate strong cross-model generalization within the evaluated setting, positioning \solpi as a preliminary step toward scalable RSI systems.

\subsection{Limitations and Future Directions}

\paragraph{Pre-Training the Harness.} Generalizing harness artifacts discovered through RSI remains a persistent challenge. Our results provide preliminary evidence that scaling auto-research loops can mitigate this challenge. Analogous to pretraining, the harness is exposed to many tasks and updated from the resulting trajectories. We hypothesize that scaling both executable environments and the diversity of research ideas can yield sustained gains; we call this long-term research direction \emph{pretraining the harness}. We will continue to explore this direction.

\paragraph{Multi-Backend Training.} Treating the harness as trainable suggests a complementary path: learning from multiple LLM backends. Our current harness was updated from trajectories generated by a single LLM; on the other evaluated backend, its mechanisms trigger less often and less intensively, although they still improve token efficiency when triggered. Training and validating the harness across multiple backends may improve robustness while preserving task quality and efficiency.

\paragraph{Recursive Efficient Improvement.} Efficiency may also become recursive: a more efficient harness could lower the cost of the auto-research used to build its successor. We plan to use \solpi as the starting harness for the next research cycle, where lower per-run costs could let a fixed budget cover more executable environments, trajectories, and research ideas. In this view, efficiency is both an outcome of harness research and a resource for expanding the search that follows, so a more efficient harness may help discover an even more efficient one. We call this possibility \emph{recursive efficient improvement}; it is a long-term research vision rather than a compounding effect demonstrated by the present study.

\paragraph{Search Coverage and Cost.} Running complete auto-research loops in our environment is computationally expensive, making controlled comparisons of search breadth and depth under a fixed budget particularly challenging. We view scaling laws along these dimensions as a promising research direction and leave their systematic investigation to future work.

{
  \small
  \bibliographystyle{unsrtnat}
  \bibliography{main}
}

\end{document}